\documentclass[]{article}
\usepackage[letterpaper]{geometry}
\usepackage{amta2022}
\usepackage{times}
\usepackage{url}
\usepackage{latexsym}
\usepackage{natbib}
\usepackage{layout}
\usepackage{amssymb}
\usepackage{tabularx}
\usepackage{amsmath}
\usepackage{graphicx}
\usepackage{booktabs,chemformula}
\definecolor{darkgreen}{rgb}{0.0, 0.5, 0.0}
\usepackage{tablefootnote}

\begin{document}

\title{Embedding-Enhanced GIZA++: \\ 
Improving Word Alignment Using Embeddings}

\author{\name{\bf Kelly Marchisio} \hfill  \addr{kmarc@jhu.edu}\\
        \addr{Department of Computer Science, Johns Hopkins University,
        Baltimore, MD, 21211, USA}
\AND
       \name{\bf Conghao Xiong\thanks{\hspace{2mm}Work completed at Johns Hopkins University.}}  \hfill \addr{chxiong21@cse.cuhk.edu.hk}\\
 \addr{Department of Computer Science and Engineering, The Chinese University of Hong Kong, New Territory, HKSAR}
\AND
       \name{\bf Philipp Koehn} \hfill \addr{phi@jhu.edu}\\
        \addr{Department of Computer Science, Johns Hopkins University,
        Baltimore, MD, 21211, USA}
}
\maketitle
\pagestyle{empty}

\begin{abstract}
A popular natural language processing task decades ago, word alignment has been dominated until recently by GIZA++, a statistical method based on the 30-year-old IBM models. New methods that outperform GIZA++ primarily rely on large machine translation models, massively multilingual language models, or supervision from GIZA++ alignments itself.  We introduce Embedding-Enhanced GIZA++, and outperform GIZA++ without any of the aforementioned factors.  Taking advantage of monolingual embedding spaces of source and target language only, we exceed GIZA++'s performance in every tested scenario for three languages pairs.  In the lowest-resource setting, we outperform GIZA++ by 8.5, 10.9, and 12 AER for Ro-En, De-En, and En-Fr, respectively. We release our code at \url{https://github.com/kellymarchisio/ee-giza}.
\end{abstract}

\section{Introduction}
Word alignment techniques were once ubiquitous in the machine translation (MT) literature, as they formed a critical part of statistical machine translation (SMT) systems.  Since the advent of neural machine translation (NMT), word alignment is no longer a step in typical NMT training, but is still important for other tasks such as annotation transfer \citep[e.g.][]{yarowsky2001inducing, rasooli2018cross}, as a post-processing step of MT to reinsert markup \citep[e.g.][]{muller-2017-treatment}, and for some mapping-based unsupervised MT methods such as \citet{artetxe-etal-2019-effective}. 

GIZA++ \citep{och-2003-minimum}, a statistical alignment model, has been the most commonly used tool for word alignment quality for 20 years and is based the IBM translation models that are yet a decade older \citep{brown-etal-1993-mathematics}.  Though a handful of neural systems have outperformed GIZA++, these rely on large MT models \citep[e.g.][]{stengel-eskin-etal-2019-discriminative,chen-etal-2020-accurate, zenkel-etal-2020-end}, massively multilingual language models \citep[e.g.][]{garg-etal-2019-jointly,jalili-sabet-etal-2020-simalign, dou-neubig-2021-word}, supervision from human-annotated alignments \citep{nagata-etal-2020-supervised}, or combinations of the above. 

We introduce Embedding-Enhanced GIZA++ (EE-GIZA++), an improvement to GIZA++ without any of the aforementioned factors. EE-GIZA++ biases GIZA++ to align semantically similar words from a shared embedding space. We outperform GIZA++ in all tested settings on three language pairs. EE-GIZA++ is particularly strong in comparison with GIZA++ when parallel training data is scarce: using only $\sim$500 lines of bitext, it outperforms GIZA++ by 10.9 AER\footnote{Alignment Error Rate (Och and Ney, 2000a).} and 12.0 AER for De-En and Fr-En, respectively. 

\section{Related Work}
Fast-align is a statistical aligner similar to GIZA++. It is a reparameterization of IBM Model 2 \citep{dyer-etal-2013-simple}. eflomal is another highly-performant non-neural aligner \citep{ostling2016efficient}. We use GIZA++ as our base system because it commonly-used and trusted for generating high-quality alignments. Numerous improvements to GIZA++ have been proposed \citep[e.g.][]{vaswani-etal-2012-smaller}.

Recent work involves using neural translation models to guide or extract alignments, viewing attention as a proxy for alignment \cite[e.g.][]{peter2017generating, li2018target, garg-etal-2019-jointly, zenkel2019adding, zenkel-etal-2020-end, chen-etal-2020-accurate}. Other aligners use massive multilingual language models with contextualized embeddings such as mBERT \citep{devlin-etal-2019-bert}. Like us, \citet{jalili-sabet-etal-2020-simalign} experiment with mapped monolingual embedding spaces, but exceed the GIZA++ baseline only when using spaces such as mBERT and XLM-R \citep{conneau-etal-2020-unsupervised}. \citet{dou-neubig-2021-word}'s approach is similar to the aforementioned authors, but they improve results by finetuning mBERT on auxiliary tasks. \citet{nagata-etal-2020-supervised} use mBERT and require supervision with human-annotated alignments.

\citet{pourdamghani-etal-2018-using} use word embedding similarity to augment parallel data seen by GIZA++, improving alignment and downstream low-resource MT. \citet{jalili-sabet-etal-2016-improving} also use nearest-neighbors in a word embedding space to alter IBM Model 1, but their performance does not match ours. Perhaps most similar to our work, \citet{songyot-chiang-2014-improving} incorporate word similarity into GIZA++ using a feedforward neural network trained to model word similarity, with a hyperparameter to control the influence of the neural model.    

\section{Background}
Let $S$ be a source-language sentence of tokens $(s_1, s_2, ..., s_m)$ and $T$ be a target-language sentence $(t_1, t_2, ..., t_l)$. Alignments are defined as $A \subseteq \{(s, t) \in S \times T\}$ where each $s, t$ are meaningfully related---usually, translations of one another. Performance is typically measured with Alignment Error Rate \citep[AER;][]{och2000improved}.

\subsection{GIZA++}
GIZA++ is a popular statistical alignment and MT toolkit \citep{och-ney-2000-improved, och-ney-2003-systematic} which implements IBM Models 1-5 \citep{brown-etal-1993-mathematics} and the HMM Model \citep{vogel1996hmm}, trained using expectation-maximization (EM). The default training setup is to run five iterations each of IBM Model 1, HMM, Model 3, and Model 4. GIZA++ is highly effective at aligning frequent words in a corpus, but error-prone for infrequent words. 
\paragraph{IBM Models}
The IBM models developed more than 30 years ago for MT are useful for alignment. IBM Model 1 relies on lexical translation probabilities $p(f|e)$ for source word $e$ and target word $f$. Model 2 adds an alignment model $p(j \mid i, l, m)$, predicting source position $j$ from target position $i$ of sentences with lengths $m$ and $l$, respectively. Model 3 adds a fertility model. Model 4 and the HMM Model replace the alignment with a relative reordering model. After training, the most likely alignment can be computed for a sentence pair.
\subsection{Monolingual Embedding Space Mapping}
Non-contextual vector representations of words (``word embeddings", ``word vectors") are common in NLP \cite[e.g.][]{mikolov2013, bojanowski2017enriching}. Word vectors trained on monolingual data \textit{embed} the word into an N-dimensional space where distance and angle have meaning. Mapping monolingual embedding spaces to a shared crosslingual space is common, particularly for bilingual lexicon induction and cross-lingual information retrieval. 
\paragraph{Procrustes Problem}
Techniques that map monolingual embedding spaces to a crosslingual space often solve a variation of the generalized Procrustes problem \cite[e.g.,][]{artetxe-etal-2018-robust, conneau-lample-2018, patra-etal-2019-bilingual, ramirez2020novel}.
Given word embedding matrices $X, Y \in \mathbb{R}^{n\times d}$ where $x \in X$, $y \in Y$ are word vectors in source and target languages, one finds the map $W \in \mathbb{R}^{d\times d}$ that minimizes distances for each pair $(x, y)$ known to be translations:
\begin{equation*}
    \setlength{\abovedisplayskip}{3pt}
    \setlength{\belowdisplayskip}{1.5pt}
\underset{W}{\arg\min}\lVert XW - Y\rVert_F
\end{equation*}
When restricting W to be orthogonal (${WW^T=I}$), \citet{schonemann1966generalized} showed that the closed-form solution is $W = VU^T$, where $U\Sigma V$ is the singular value decomposition of $Y^TX$.

After mapping $X$ and $Y$ to a shared space with $W$, translations are extracted via nearest-neighbor search. A popular distance metric is cross-domain similarity local scaling (CSLS) to mitigate the ``hubness problem" \citep{conneau-lample-2018}. 

\section{Method}
\begin{figure*}[htb]
  \centering
  \includegraphics[width=1\linewidth]{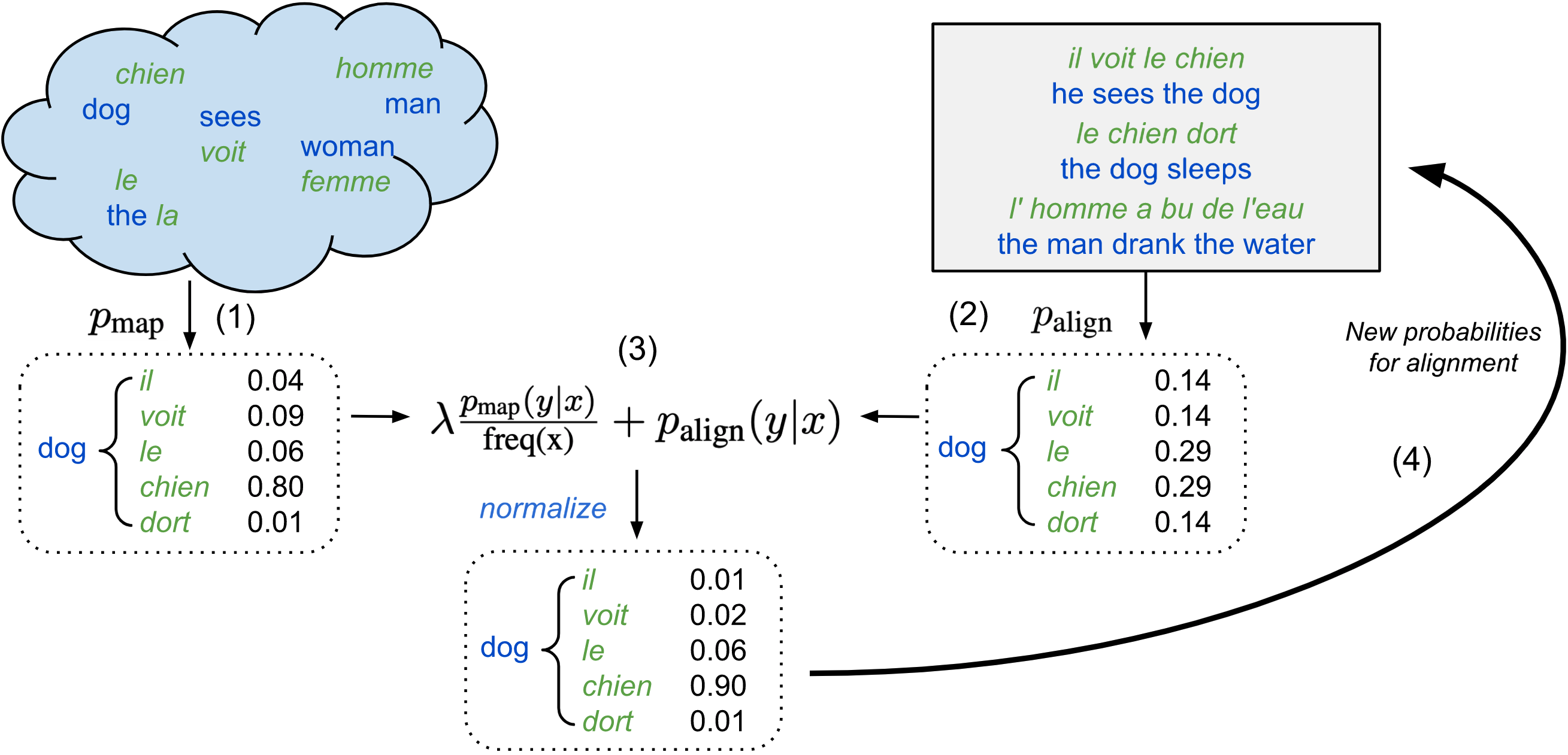}
  \caption{Proposed Method: Embedding-Enhanced GIZA++. 1) Map monolingual embeddings to crosslingual space. Calculate CSLS for cooccurring words and take softmax to calculate a probability distribution (\textit{p\_map}). 2) Use statistical aligner to calculate separate probability distribution over cooccuring words (\textit{p\_align}). 3) Interpolate distributions with weight proportional to source word's frequency. Normalize. 4) Replace the statistical model's translation probability table with updated probability distribution. 5) Repeat Steps 2-4 for each iteration of EM.}
  \label{fig:method}
\end{figure*}
GIZA++ is highly effective at inducing the correct alignment for frequent words when parallel resources are abundant, but is error-prone for rare words. Because word embeddings can be trained on large amounts of monolingual data, rare words from a parallel corpus may be well-enough represented in a large monolingual corpus that reasonable word embeddings can be trained. Our key insight is that for infrequent words, finding a translation via nearest-neighbors in a shared embedding space may be more reliable than using a statistical aligner. We thus incorporate embedding space mapping into GIZA++ training, giving more or less influence to the statistical aligner depending on word frequency. Figure \ref{fig:method} shows the method.
\paragraph{1. Map embedding spaces.}
Word embedding spaces $X$ and $Y$ for source and target language, respectively, are mapped to a crosslingual space using VecMap\footnote{\url{github.com/artetxem/vecmap}} \citep{artetxe2018generalizing}.   

\paragraph{2. Calculate translation probability distribution from mapped spaces.}
\label{geom-probdist}
Let $\text{Co}_{Y}(x)$ be the words from the target language that cooccur with source word $x$ in the corpus.  
For each $x$, we calculate a probability distribution over possible alignments from $\text{Co}_{Y}(x)$ with a softmax over the CSLS scores (We use $\tau = 0.1$.) We use the mapped embedding spaces for source and target languages to calculate CSLS.
\begin{equation*}
p_{map}(y|x) = \frac{\exp{(\text{CSLS}(x,y)/\tau)}}{\sum\limits_{y' \in \text{Co}_{Y}(x)}{\exp{(\text{CSLS}(x,y'))/\tau)}}}
\end{equation*}
\paragraph{3. Integrate with GIZA++.}
\label{integration}
Recall that IBM Models 1, 3, 4, and HMM maintain a lexical translation table of $p_{\text{align}}(y\vert x)$ for every cooccurring source-target word pair.
During training of IBM Model 1 and the HMM, we interpolate the lexical translation table with embedding-based translation probabilities after each iteration of EM. For each cooccurring pair $(x,y)$, calculate:
\begin{equation*}
score(x, y) = \lambda \frac{ p_\text{map}(y\vert x)}{\text{freq(x)}} + p_\text{align}(y\vert x)
\end{equation*}

\noindent where freq(x) is the raw frequency of $x$ in the source-side of the corpus and $\lambda$ is a hyperparameter.  The effect is that $p_{map}$ is given more weight for infrequent words, in accordance with our goal to trust the embedding space mapper for infrequent words and the statistical aligner for frequent words.  Then normalize over cooccuring words:
\begin{equation}
\label{eq:new_palign}
p(y\vert x) = \frac{score(x, y)}
{\sum\limits_{y_i\in  \text{Co}_{Y}(x)} score(x, y_i)}  
\end{equation}

We update GIZA++'s lexical translation table with the new value from Equation \ref{eq:new_palign} for all cooccurring pairs, then begin the next iteration of EM.\footnote{If a word from the bitext is not present in the word embedding space, its translation probability is not updated.} This process is repeated for all iterations of IBM Model 1 and HMM model training. IBM Model 3 and 4 are trained as usual. Integrating probabilites from $p_{map}$ into IBM Models 3 and 4 is for future work. 
\paragraph{}
Steps 1-3 are done in source$\rightarrow$target and target$\rightarrow$source directions. Alignments are symmetrized with grow-diag-final \citep{koehn-etal-2003-statistical}.

\section{Experimental Setup}
We use the same training setup as previous work\footnote{\url{https://github.com/lilt/alignment-scripts}. Data: \citep{mihalcea2003evaluation, koehn2005europarl, vilar2006aer}} \citep{garg-etal-2019-jointly, zenkel2019adding, zenkel-etal-2020-end, chen-etal-2020-accurate, dou-neubig-2021-word}. Training corpora for German-English (De-En), English-French (En-Fr), and Romanian-English (Ro-En) are 1.9M, 1.1M, and 448K lines, and test sets are 508, 447, and 248 lines, respectively. Validation sets do not exist, so we tune $\lambda$ on 1 million lines of De-En.\footnote{This was the approximate average size of training data for all languages.} $\lambda$ is set to 10,000. We use the VecMap implementation of CSLS and SciPy for some utility functions and softmax calculation \citep{2020SciPy-NMeth, 2020NumPy-Array}. For pretrained word embedding spaces, we use the publicly-available Wikipedia word vectors trained using fastText from \citet{bojanowski2017enriching}.\footnote{\url{https://fasttext.cc/docs/en/pretrained-vectors.html}} We limit vocabulary size to 200,000 and perform embedding mapping with VecMap in unsupervised mode.

\begin{table*}[ht]
  \centering
  \begin{tabular}{@{}r|cc|cc|cc@{}}
    \toprule
 &  \multicolumn{2}{c}{\underline{De-En}} & \multicolumn{2}{c}{\underline{Ro-En}} & \multicolumn{2}{c}{\underline{En-Fr}} \\
Corpus Size & 	GIZA++	&	Ours & 	GIZA++	&	Ours & 	GIZA++	&	Ours \\ 
  \midrule
Test Set Only &	44.2	&	\textbf{33.3} \textcolor{darkgreen}{\textit{(-10.9)}}	&	42.8 & \textbf{34.3} \textcolor{darkgreen}{\textit{(-8.5)}} & 26.9 & \textbf{14.9} \textcolor{darkgreen}{\textit{(-12.0)}}      \\
1000    & 41.0      &  \textbf{31.1}	 \textcolor{darkgreen}{\textit{(-9.9)}} & 41.5 & \textbf{33.6} \textcolor{darkgreen}{\textit{(-7.9)}} & 20.0 & \textbf{11.4} \textcolor{darkgreen}{\textit{(-8.6)}} \\
2000    & 37.7      &  \textbf{29.1}	 \textcolor{darkgreen}{\textit{(-8.6)}} & 39.6 & \textbf{32.9} \textcolor{darkgreen}{\textit{(-6.7)}} & 17.2 & \textbf{10.1} \textcolor{darkgreen}{\textit{(-7.1)}} \\
5000    & 34.5      &  \textbf{26.9}	 \textcolor{darkgreen}{\textit{(-7.6)}} & 38.2 & \textbf{32.0} \textcolor{darkgreen}{\textit{(-6.2)}} & 14.0 & \textbf{8.5} \textcolor{darkgreen}{\textit{(-5.5)}} \\
10,000   & 31.9      &  \textbf{25.5}	 \textcolor{darkgreen}{\textit{(-6.4)}} & 36.1 & \textbf{30.4} \textcolor{darkgreen}{\textit{(-5.7)}} & 11.7 & \textbf{7.5} \textcolor{darkgreen}{\textit{(-4.2)}} \\
20,000   & 29.3      &  \textbf{24.2}	 \textcolor{darkgreen}{\textit{(-5.1)}} & 35.2 & \textbf{30.3} \textcolor{darkgreen}{\textit{(-4.9)}} & 10.0 & \textbf{7.1} \textcolor{darkgreen}{\textit{(-2.9)}} \\
50,000   & 26.6      &  \textbf{22.6}	 \textcolor{darkgreen}{\textit{(-4.0)}} & 34.2 & \textbf{29.7} \textcolor{darkgreen}{\textit{(-4.5)}} & 8.6 & \textbf{6.3} \textcolor{darkgreen}{\textit{(-2.3)}} \\
100,000  & 25.4      &  \textbf{21.9}	 \textcolor{darkgreen}{\textit{(-3.5)}} & 33.4 & \textbf{29.3} \textcolor{darkgreen}{\textit{(-4.1)}} & 7.8 & \textbf{6.1} \textcolor{darkgreen}{\textit{(-1.7)}}  \\
200,000  & 24.0      &  \textbf{21.2}	 \textcolor{darkgreen}{\textit{(-2.8)}} & 32.7 & \textbf{29.4} \textcolor{darkgreen}{\textit{(-3.3)}} & 7.0 & \textbf{5.8} \textcolor{darkgreen}{\textit{(-1.2)}}  \\
500,000  & 21.6      &  \textbf{20.3}	 \textcolor{darkgreen}{\textit{(-1.3)}} & 26.5 & \textbf{25.5} \textcolor{darkgreen}{\textit{(-1.0)}} & 6.1 & \textbf{5.7} \textcolor{darkgreen}{\textit{(-0.4)}}  \\
1,000,000	&	20.7	&	\textbf{20.1}    \textcolor{darkgreen}{\textit{(-0.6)}} & \textit{n/a} & \textit{n/a} & 6.1 & \textbf{5.5} \textcolor{darkgreen}{\textit{(-0.6)}}  \\
1,900,000	&	20.6	&	\textbf{19.9}    \textcolor{darkgreen}{\textit{(-0.7)}} & \textit{n/a} & \textit{n/a} & \textit{n/a} & \textit{n/a} \\
    \bottomrule
  \end{tabular}
  \caption{Main Results. Alignment Error Rate (AER) of EE-GIZA++ vs. GIZA++ baseline (lower is better). Test set is included in corpus size. Ro-En 500K is the full 448K training set. Bidirectional, symmetrized (grow-diag-final).}
  \label{tab:main-results}
\end{table*}

\begin{figure}[h]
  \centering
  \includegraphics[height=0.27\textheight,width=0.70\linewidth]{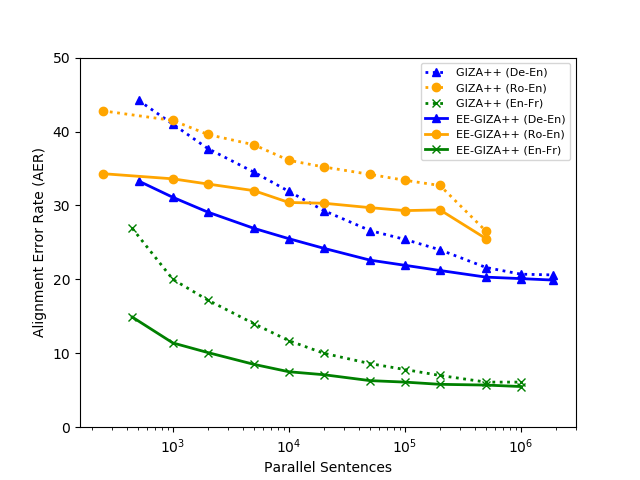}
  \caption{Visualization of Main Results. Alignment Error Rate (AER) of EE-GIZA++ vs. GIZA++ baseline for increasing amounts of training data. Lower is better.}
  \label{fig:vgiza}
\end{figure}

\section{Results}

The main results are presented in Table \ref{tab:main-results} and visualized in Figure \ref{fig:vgiza}. We observe that EE-GIZA++ consistently outperforms GIZA++ by a large margin in every tested scenario. When aligning the test set alone with no additional bitext, EE-GIZA++ dramatically outperforms GIZA++: by 8.5 AER for Ro-En, 10.9 AER for De-En, and 12 AER for En-Fr.  This represents improvements of approximately 20\%, 25\%, and 45\% for Ro-En, De-En, and En-Fr, respectively.  The error-rate improvement is especially notable when we consider that each test set has only approximately 250-500 lines.  When expanding the training set to include a total of 10,000 lines, we continue to observe strong gains with our method: with absolute improvements of 5.7, 6.4, and 4.2 AER for Ro-En, De-En, and En-Fr. These represent improvements of approximately 15.8\%, 20.1\%, and 35.9\%, respectively. 

\paragraph{Supplemental Results: High-Resource}
\begin{table}[htb]
  \centering 
  \begin{tabular}{@{}l@{ }c@{ }c@{ }c@{}}
    \toprule
    \midrule
\textit{Statistical Baselines} & \textbf{De-En} & \textbf{Ro-En} & \textbf{En-Fr}\\
\hspace{2mm}GIZA++ & 20.6 & 26.5 & 6.2 \\
\hspace{2mm}eflomal* & 22.6 & 25.1 & 8.2 \\
\hspace{2mm}fast-align* & 27.0 & 32.1 & 10.5 \\
\midrule
\textit{Massively-Multilingual} \\
\hspace{2mm}\citet{jalili-sabet-etal-2020-simalign} & 19.$\dagger$ & 27.2*\tablefootnote{As \citet{jalili-sabet-etal-2020-simalign} use the 2005 Ro-En test set from \url{https://web.eecs.umich.edu/~mihalcea/wpt05}, we report \citet{dou-neubig-2021-word}'s Ro-En results here for consistency with the others, which use the 2003 test set (\url{https://web.eecs.umich.edu/~mihalcea/wpt}.} & 6.$\dagger$\\
\hspace{2mm}\citet{dou-neubig-2021-word} & 15.6 & 23.0 & 4.4 \\
\hspace{7mm}no fine-tuning & 17.4 & 27.9 & 5.6 \\
\midrule
\textit{Bilingual NMT-Based} \\
\hspace{2mm}\citet{zenkel2019adding} & 21.2 & 27.6 & 10.0 \\
\hspace{2mm}\citet{garg-etal-2019-jointly} & 20.2 & 26.0 & 7.7 \\
\hspace{7mm} using GIZA++ output & 16.0 & 23.1 & 4.6 \\
\hspace{2mm}\citet{zenkel-etal-2020-end} & 16.3 & 23.4 & 5.0 \\
\hspace{2mm}\citet{chen-etal-2020-accurate} & 15.4 & 21.2 & 4.7 \\
\midrule
Ours & 19.9 & 25.5 & 5.3 \\
\midrule
    \bottomrule
  \end{tabular}
  \caption{Supplemental results in high-resource settings compared to models that use additional resources. ``Massively multilingual" models use mBERT. NMT models likely fail in low-bitext scenarios (our focus). Bidirectional. *reported in \protect\citet{dou-neubig-2021-word}.
  $\dagger$\citet{jalili-sabet-etal-2020-simalign} report one less significant digit.}
  \label{full-dataset-results}
\end{table}
We use the full data sets for De-En, Ro-En, and En-Fr and compare to existing work in Table \ref{full-dataset-results}.\footnote{Many of these use the grow-diag symmetrization heuristic, but we use grow-diag-final.} We outperform the three statistical baselines, except eflomal on Ro-En. EE-GIZA++ outperforms \citet{jalili-sabet-etal-2020-simalign} on Ro-En and En-Fr, which utilizes a massively-multilingual language model. \citet{dou-neubig-2021-word} with fine-tuning outperforms our model, though they use mBERT which is trained on 104 languages. Notably, \citet{garg-etal-2019-learning} use GIZA++ output as supervision. EE-GIZA++ performs better than GIZA++, so AER might improve if supervised with our alignments.  

\section{Conclusion and Future Work}
We introduce EE-GIZA++, an unsupervised enhancement to GIZA++ that uses word embeddings for improved word alignment in low-bitext settings, without the use of NMT or massively-multilingual language models that to-date have been the strongest competitors to GIZA++. EE-GIZA++ outperforms GIZA++ by 8.5, 10.9, and 12 AER in lowest-bitext scenarios for Ro-En, De-En, and En-Fr, respectively. Future work should examine performance of EE-GIZA++ on a diverse set of languages with varying scripts and amounts of data available.

\section*{Acknowledgements}
The authors would like to thank Elias Stengel-Eskin, Mahsa Yarmohammadi, and Marc Marone for fruitful discussion on alignment. The first author would like to thank Martin St. Denis for his ideas about the graphic used in this work.  

\bibliographystyle{apalike}
\bibliography{anthology, amta2022}

\end{document}